\documentclass[letterpaper, 10 pt, conference]{ieeeconf}
\usepackage{amsmath,amssymb,euscript,yfonts,psfrag,latexsym,dsfont,graphicx}
\usepackage{bbm,color,amstext,wasysym,parskip,balance}
\usepackage{caption}
\usepackage{subcaption}

\usepackage{multirow}

\usepackage{graphics} 
\usepackage{times} 
\usepackage{cite}

\usepackage{lipsum}

\graphicspath{{./},{./figures/}}

\usepackage{algorithm}
\usepackage{algorithmic}

\newcommand{\mE}{{\mathbb E}}

\newcommand{\mR}{{\mathbb R}}

\newcommand{\cN}{{\mathcal N}}

\newcommand{\cX}{{\mathcal X}}

\usepackage{hyperref}
\hypersetup{
    colorlinks=true,
    linkcolor=blue,
    filecolor=magenta,      
    urlcolor=cyan,
    pdftitle={Overleaf Example},
    pdfpagemode=FullScreen,
}

\definecolor{grey}{rgb}{0.6,0.6,0.6}
\definecolor{lightgray}{rgb}{0.97,.99,0.99}
\newlength{\bibitemsep}
\newlength{\bibparskip}
\let\oldthebibliography\thebibliography
\renewcommand\thebibliography[1]{%
  \oldthebibliography{#1}%
  \setlength{\parskip}{\bibitemsep}%
  \setlength{\itemsep}{\bibparskip}%
}

\begin{document}
\title{Efficient Belief Road Map for Planning Under Uncertainty}
\author{Zhenyang Chen, Hongzhe Yu and Yongxin Chen
\thanks{Z. Chen, H. Yu and Y. Chen are with the Georgia Institute of Technology, Atlanta, GA, USA. {\tt\small \{zchen927, hyu419, yongchen\}@gatech.edu}}}

\maketitle

\begin{abstract}
Robotic systems, particularly in demanding environments like narrow corridors or disaster zones, often grapple with imperfect state estimation. Addressing this challenge requires a trajectory plan that not only navigates these restrictive spaces but also manages the inherent uncertainty of the system. We present a novel approach for graph-based belief space planning via the use of an efficient covariance control algorithm. By adaptively steering state statistics via output state feedback, we efficiently craft a belief roadmap characterized by nodes with controlled uncertainty and edges representing collision-free mean trajectories. The roadmap's structured design then paves the way for precise path searches that balance control costs and uncertainty considerations. Our numerical experiments affirm the efficacy and advantage of our method in different motion planning tasks. Our open-source implementation can be found at \textit{https://github.com/hzyu17/VIMP/tree/BRM}.

\end{abstract}

\section{Introduction}
In the challenging realm of robotic motion planning, uncertainty presents a critical hurdle for effective operation in dynamic and complex real-world environments. Historically, motion planning under uncertainty evolved from deterministic motion planning foundations \cite{lavalle2006}, adopting one of two primary trajectories: the optimization-based approach and the sampling-based strategy.

The trajectory optimization paradigm, extensively studied in works like \cite{ratliff2009} and \cite{schulman2014}, transforms planning challenges into optimal control problems. This transformation necessitates the resolution of the Hamilton–Jacobi–Bellman equation through dynamic programming techniques. However, while this method promises precision, it faces significant scalability issues \cite{lewis2012optimal} \cite{yu2022data}, often at the cost of local solutions or even infeasibility.
\begin{figure}[tb] 
    \centering
    \includegraphics[width=0.35\textwidth]{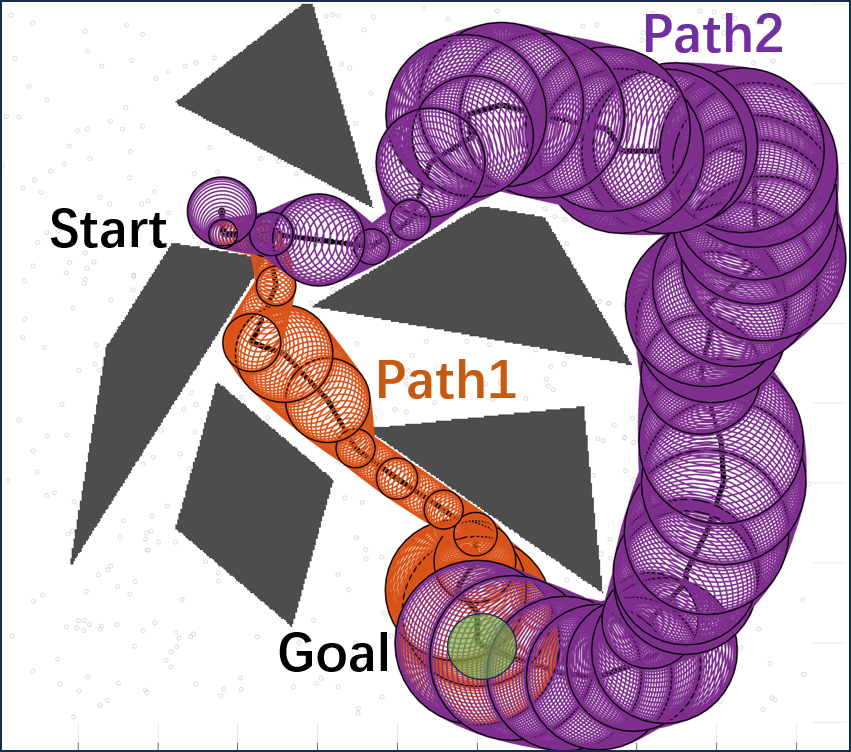}
    \caption{A belief space graph depicting sampled state beliefs and the connected edges against obstacles. When prioritizing entropy cost over control energy, the planner opts for Path 2 (shown in purple), which allows for increased uncertainty, enhancing robot safety. Conversely, if control energy minimization is paramount, the planner selects the more direct Path 1 (shown in orange), demonstrating reduced uncertainty tolerance. }
    \label{fig:BRM_two_paths}
\end{figure}
On the other end of the spectrum, sampling-based planning establishes motion planning as a search problem. By utilizing algorithms such as the probabilistic road maps (PRM) \cite{kavraki1996} and rapidly exploring random trees (RRT) \cite{lavalle1998}\cite{sertac-rrt}, this approach leverages graph structures filled with random feasible states to pinpoint optimal paths. What sets this approach apart is its promise of probabilistically complete solutions, assuring an increasing likelihood of finding a feasible solution with more samples.

While deterministic motion planning offers a robust framework, introducing uncertainties complicates the scenario significantly. This led to the development of belief space planning \cite{platt2010belief} \cite{vitus2011closed} \cite{macindoe2012pomcop} \cite{pedram_gaussian_2022}. Essentially an expansion of traditional planning to incorporate uncertainties, this approach has seen a growing emphasis on belief road maps (BRM) \cite{prentice2009belief} \cite{shan2017belief} \cite{wang2018efficient} \cite{zheng_belief_2021}. Unlike the conventional nodes of deterministic states in PRM, BRM employs state distributions, bringing forth unique challenges, especially regarding computational efficiency.

At the intersection of these challenges lies covariance steering, a discipline geared towards guiding distributions. Notably, in a series of studies \cite{CheGeoPav15a} \cite{CheGeoPav15b}, Chen et al. illustrated that linear system distribution steering can be achieved through closed-form solutions. More recent works \cite{yu2023stochastic} \cite{yu2023covariance} expanded on this by integrating safety constraints and addressing nonlinear control-affine dynamics.

Building upon these advances, our research introduces a nuanced covariance steering approach for graph-based motion planning. We tackle the BRM's existing challenges by proficiently crafting probabilistic graph edges. Incorporating state estimation \cite{chen2015steering} further aligns our methodology with the broader belief space planning framework, drawing parallels to chance-constrained strategies like CC-RRT* \cite{Luders2013RobustSM}. Empirical evidence, as we will present, accentuates the advantages of our approach over existing methods, showcasing both its effectiveness and efficiency in managing uncertainties.

\section{Background}\label{sec:back}
This section introduces belief space planning and covariance steering as key components of our proposed method.

\subsection{Belief space planning}
Belief space planning addresses the challenge of making decisions with uncertain robot states where the belief state $b$ is a composite representation of the robot's state and its associated uncertainty. With new input $u$ and observation $z$, the state transition function $\tau$ updates the belief state $b^{'}=\tau(b,u,z)$. Instead of always choosing the shortest path, belief space planners leverage the belief information and search for a more conservative motion plan when the state estimation is uncertain, as shown in figure \ref{fig:BRM_two_paths}. A significant concern when planning in belief spaces is the computational challenge due to the high dimensionality of belief states. This problem can be addressed by sampling-based algorithms like PRM. 

The belief-space variant of the PRM is called the Belief Roadmap (BRM) \cite{prentice2009belief}. The primary idea of BRM is to sample both configurations and their distributions in the belief state space, test them for feasibility, and then attempt to connect nearby configurations to form a roadmap. The BRM can be mathematically represented as a graph \( G = (V, E) \) where \( V=\{b_i\} \) is the set of nodes representing feasible belief states and \( E \) is the set of edges indicating belief paths between adjacent nodes. To construct BRM, for each pair of belief nodes \( (b_i, b_j) \), a local planner attempts to find a feasible path considering both the spatial constraints and the belief evolution. The belief evolution accounts for uncertainty propagation, influenced by robot dynamics and environmental factors. Once the belief roadmap is constructed with the cost associated with traversal and belief uncertainty, an optimal path can be found by graph search algorithms like $A^*$ and Dijkstra.

\subsection{Covariance steering for control-affine systems}\label{sec:linearcov}
The covariance steering problem for nonlinear systems remains a challenge. Recent progress established in \cite{yu2023covariance} demonstrates an efficient algorithm tailored for control-affine systems. We present the main results in this section. The nonlinear system under consideration is
	\begin{equation}\label{eq:nonlinear}
		dX_t = f(t,X_t)dt + B(t) (u_t dt + \sqrt{\epsilon} d W_t)
	\end{equation}
where $X_t\in \mR^n$ is the state vector, $u_t\in \mR^p$ is the input vector and $f(t,X_t)$ is the drift function. The input matrix $B(t)\in\mR^{n\times p}$ is assumed to be full rank. $W_t\in \mR^p$ represents a standard Wiener process \cite{CohEll15}, and $\epsilon>0$ parameterizes the intensity of the disturbance. The covariance steering problem minimizes the control energy while seeking a state feedback policy to steer state statistics of the system from an initial value to a terminal one. 
	\begin{subequations}\label{eq:covcontrol}
	\begin{eqnarray}\label{eq:covcontrol1}
		\min_{u} && \mE \left\{\int_0^T [\frac{1}{2}\|u_t\|^2 + V(X_t)]dt\right\}
		\\\label{eq:covcontrol2}
		&&dX_t = f(t,X_t)dt + B(t) (u_t dt + \sqrt{\epsilon} d W_t)
		\\\label{eq:covcontrol3}
		&& 
		X_0 \sim \rho_0,\quad X_T \sim \rho_T,
	\end{eqnarray}
	\end{subequations}
where $\rho_0$ ($\rho_T$) is a probability distribution with mean $m_0$ ($m_T$) and covariance $\Sigma_0$ ($\Sigma_T$).

By leveraging the Girsanov theorem, problem \eqref{eq:covcontrol} can be transferred into a composite optimization problem which can be solved by the proximal gradient algorithm. From the results established in \cite{yu2023covariance}, each proximal gradient iteration with step size $\eta$ amounts to solving the following linear covariance steering problem
	\begin{subequations}\label{eq:iterlinear}
	\begin{eqnarray}\label{eq:iterlinear1}
		\hspace{-0.4cm}\min_{u} && \!\!\!\! \!\!\!\!  \mE \left\{\int_0^T [\frac{1}{2}\|u_t\|^2 + \frac{1}{2} X_t^T Q_k(t) X_t + X_t^T r_k(t)]dt\right\}
		\\ 
        \label{eq:iterlinear2}\!\!&& \!\!\!\! \hspace{-0.7cm}dX_t = \frac{1}{1+\eta}[ A_k(t)+\eta\hat A_k(t)] X_t dt + \frac{1}{1+\eta}[a_k(t)
		\\\nonumber \!\!\!\!&& \!\!\!\!\!\! \hspace{0.8cm} + \eta\hat a_k(t)] dt + B(t)(u_tdt+ \sqrt{\epsilon} dW_t)
		\\ \label{eq:iterlinear3}&&\!\!\!\! X_0 \sim \rho_0,\quad X_T \sim \rho_T,
	\end{eqnarray}
	\end{subequations} 
where $A_k(t),a_k(t)$ are the results from last iteration. Also, $\Bar{x}_k(t)$ is the mean trajectory at $k$, $\hat A_k(t)$, $\hat a_k(t)$ are linearization matrices along $\Bar{x}_k(t)$, and $Q_k(t),r_k(t)$ are the weighting matrices \cite{yu2023covariance}.

This result bridges the gap between the non-linear covariance steering problem and the linear covariance steering problem. 
The linear covariance steering problem in \ref{eq:iterlinear} enjoys a closed-form feedback solution in the form \cite{CheGeoPav15a} \cite{CheGeoPav17a}
\[
u_t^\star = -B(t)^T\Pi(t) (X_t-x_t^\star) + v_t^\star.
\]
 where $\Pi(t)$ satisfies a coupled Riccati equations. This closed-form solution for the proximal gradient update allows us to solve the covariance steering problem for the control-affine systems with a sublinear rate \cite{yu2023covariance}.

\section{Problem formulation}\label{sec:formulation}
In this work, we consider the motion planning problem under uncertainty. Uncertainty of the robot results from three sources: robot motion, robot state estimation, and environment. In our work, we assume the environment is deterministic and only considers the uncertainty of the robot itself. Robots are nonlinear control-affine systems whose dynamics and sensor models are
        \begin{subequations} \label{eq:robotdynamic}
            \begin{align}
                  dX_t &= f(t,X_t) dt + B(t)(u_tdt+ \sqrt{\epsilon} dW_t) 
                 \\z(t)&= h(X_t)+v(t), \quad v(t) \sim \cN(0, R(t))
            \end{align}
        \end{subequations}
Here, the notations follow the above Section \ref{eq:nonlinear} and $z(t)$ is the observation output with function $h$ and Gaussian noise $v(t)$. The dynamics and sensor model can be viewed as the belief transition function of the robot. For the uncertainty that stems from the robot motion and dynamic model, we denote $\Sigma$ as the covariance of the actual robot states $x$, which follows \ref{eq:robotdynamic}. $\Sigma$ describes the influence of noise $W_t$ on the ideal robot states which follow the uncorrupted dynamic model. 

For the uncertainty that stems from the state estimation, we denote $P(t)$ as the state error-covariance of the estimation error $\Tilde{x}$. It is worth noting that the covariance of the terminal state is required to be larger than the state error-covariance $\Sigma_T > P(T)$ when using state output as feedback \cite{chen2015steering}. Denote estimated robot states as $\hat x= x - \Tilde{x}$ and its covariance as $\hat \Sigma = \Sigma - P$. We hope to steer the state covariance $\Sigma$ by controlling the estimated state covariance $\hat \Sigma$. We can state our control problem as for the given waypoints $x_0,x_T$ and their estimated state covariance $\hat\Sigma_0, \hat\Sigma_T$, finding a control sequence $u_t$ such that 1) control the mean of the robot states from $x_0$ to $x_T$, 2) control the covariance of the robot states from $\Sigma_0$ to terminal covariance $\Sigma_T$ via output feedback $\hat x$, 3) generate a collision-free mean trajectory and 4) minimize the objective function of expected control energy and a state cost
	\begin{equation}\label{eq:mpproblem}
		\min_{u} \mE \left\{\int_0^T [\frac{1}{2}\|u_t\|^2 + V(X_t)]dt\right\}
	\end{equation}
\section{belief space collision-avoiding covariance steering}\label{sec:algorithm}
We hope to build a BRM and solve problem \eqref{eq:mpproblem} by edge construction and graph search. Constructing edges in belief space is challenging in terms of computation \cite{prentice2009belief} since it involves steering state statistics under safety constraints using partially observable state information. We leverage the proximal gradient algorithm for problem \eqref{eq:covcontrol} with a collision-avoiding state cost \cite{yu2023stochastic}
\begin{equation}
    V(X) = \lVert {\rm hinge}(S(X)) \rVert^2
    \label{eq:hinge_loss}
\end{equation}
to achieve the node connection in a BRM. In \eqref{eq:hinge_loss}, ${\rm hinge}(\cdot)$ represents the hinge loss function, and $S(\cdot)$ is a differentiable signed distance function to the obstacles. We showed in \cite{yu2023stochastic} that the proposed proximal gradient algorithm in \cite{yu2023covariance} is effective and efficient in producing collision-free belief space trajectories.

\subsection{Collision avoiding covariance steering}
\begin{figure*} \label{fig:vis_pgcs}
\centering
    \begin{subfigure}[b]{0.4\textwidth}
        \centering
    \includegraphics[width=\textwidth]{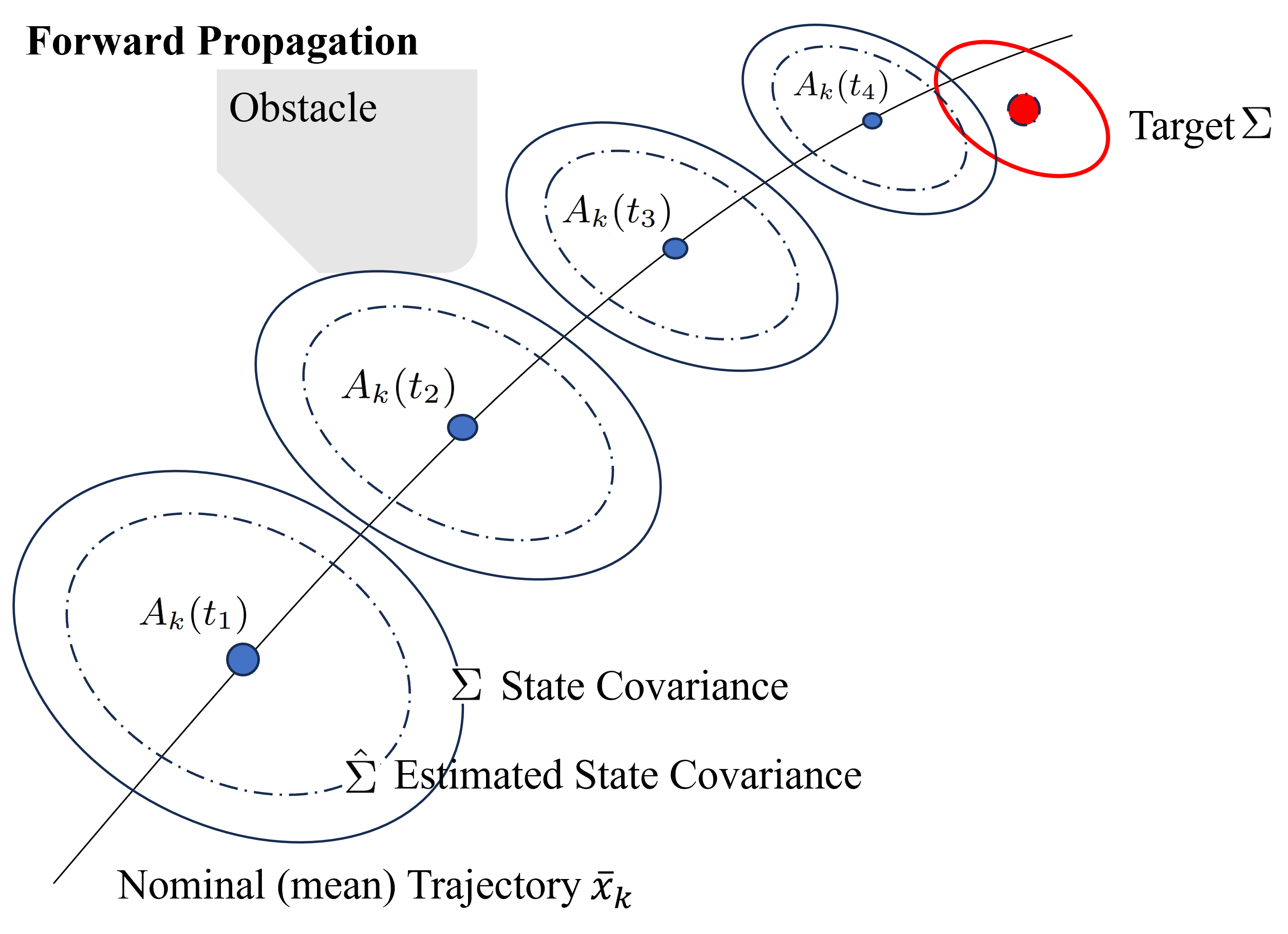}
    \caption{Forward propagation.}
    \label{fig:forward}
    \end{subfigure}
    \;\;\;\;\;\;\;\;\;\;
    \begin{subfigure}[b]{0.4\textwidth}
        \centering
    \includegraphics[width=\textwidth]{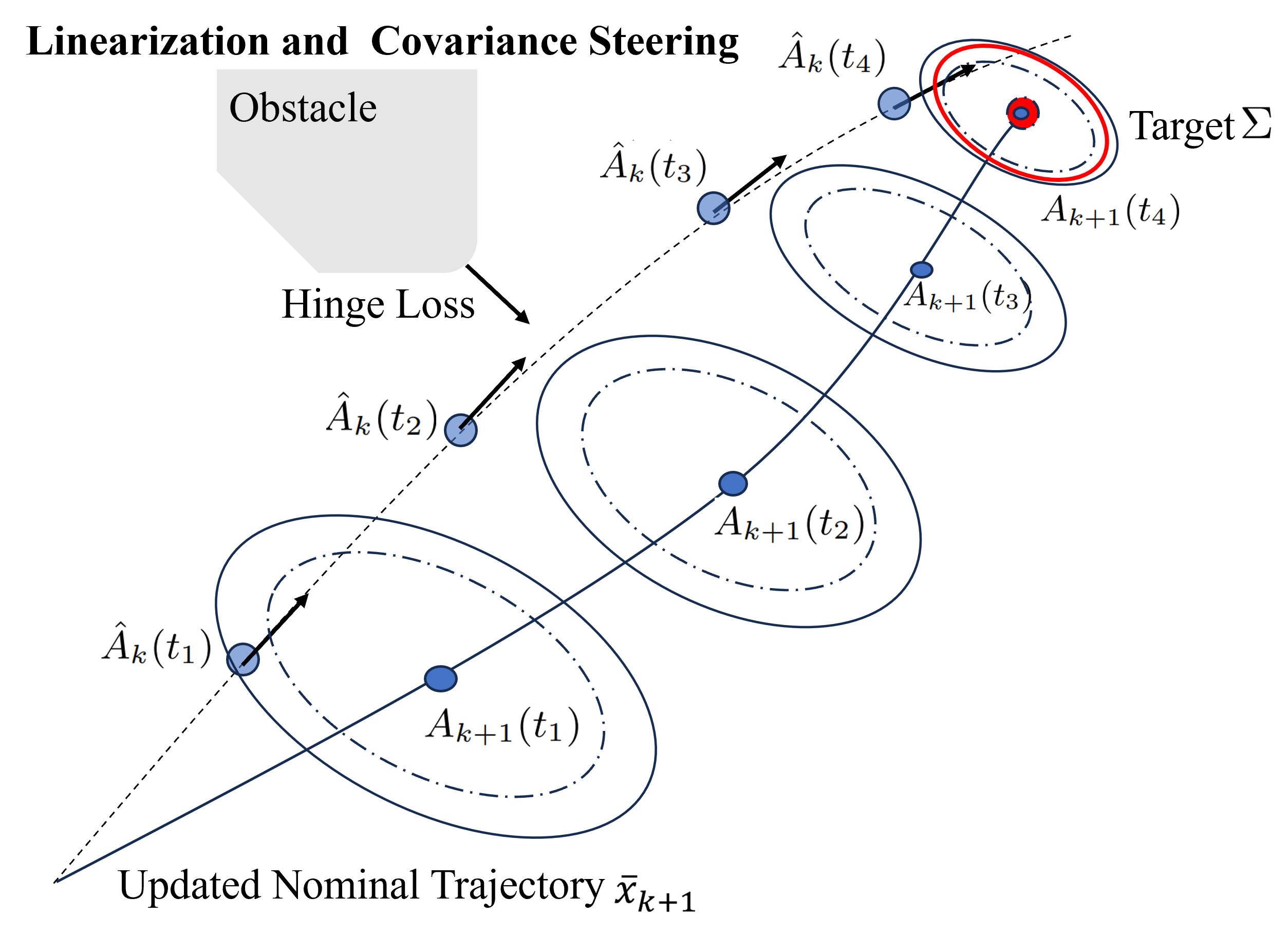}
    \caption{Collision-avoiding covariance steering.}
    \label{fig:backward}
    \end{subfigure}
    \caption{Procedure of edge construction in PGCS-BRM.}
    \label{fig:forward_backward}
\end{figure*}

By employing the hinge loss function \( {\rm hinge}(\cdot) \), we can succinctly define our cost function as \eqref{eq:hinge_loss} to penalize risky behaviors and circumvent obstacle collisions. For each iteration of the proximal gradient covariance steering algorithm associated with \eqref{eq:iterlinear}, rather than detailing the intricate mathematics of deriving the weighting matrices \( Q_k(t) \) and \( r_k(t) \), it suffices to say that they are formulated based on the gradient and Hessian of the cost function. They integrate the effects of system dynamics, control inputs, and uncertainties.

Upon solving \eqref{eq:iterlinear} within the paradigm of linear covariance steering, the optimal control policy is formulated as:
\[ u_t^\star = K_k(t) X_t + d_k(t) \]
This control policy, when injected into the closed-loop process, offers the subsequent update dynamics
\begin{align*}
dX_t &= \frac{1}{1+\eta}[A_k(t)+\eta \hat A_k(t)] X_t dt +\\&  \frac{1}{1+\eta}[ a_k(t)+\eta\hat a_k(t)] dt 
+ B(t)(u_t^\star dt+ \sqrt{\epsilon} dW_t),
\end{align*}
From this, we deduce the iterative update rules:
\begin{subequations}\label{eq:updateAa}
\begin{eqnarray}\label{eq:updateAa1}
A_{k+1}(t) &=&  \frac{1}{1+\eta}[ A_k(t)+\eta\hat A_k(t)] + B(t)K_k(t)
\\\label{eq:updateAa2}
a_{k+1}(t) &=& \frac{1}{1+\eta}[a_k(t)+\eta\hat a_k(t)] + B(t)d_k(t).
\end{eqnarray}
\end{subequations}
To synchronize the evolution of \(\Bar{x}_k(t)\) and \(\Sigma_k(t)\) at each iteration \( k \), one can employ the aforementioned update rule \eqref{eq:updateAa}, ensuring an efficient iterative process.

In the following discourse, we showcase the state connection algorithm (as presented in Algorithm \ref{alg:noncov}). Given the constructs \( A_k(t) \) and \( a_k(t) \) at the \( k^{th} \) iteration, the algorithm commences by propagating the mean trajectory \(\Bar{x}_k(t)\) and subsequently estimating the state covariance along the path, as represented in Figure \ref{fig:forward}. Leveraging the updated nominal trajectory, the algorithm exploits the control-observation separation principle to compute both the Kalman gain and the state error-covariance \( P_k(t) \).

\subsection{Steering state statistics using partially-observed output}
To initialize the state prediction for each sampled state, we set $\hat{x}_k(t_0)=\mE[x_k(t_0)]$ and $P_k(t_0)$ is sampled from a proper space. At each iteration, the continuous-time EKF propagates state error covariance $P_k(t)$ based on the linearized system dynamics model $A_k(t), a_k(t)$ and updates the near-optimal Kalman gain. These steps are coupled in continuous time and governed by the following Riccati equations
    \begin{equation} \label{eq:KFriccati}
    \begin{split}
        \dot P(t) & = F(t)P(t)+P(t)F(t)^T+B(t)QB^T(t)\\
                  & -P(t)H(t)^TR(t)^{-1}H(t)P(t)
    \end{split}
    \end{equation}
    where noise covariance $Q=\epsilon \bold{I}_n$ and $F(t)$ and $H(t)$ represent the Jacobian matrices of the system dynamics function and measurement function, respectively, as
\begin{align*}
F(t) = \frac{\partial f}{\partial x}\bigg|_{\hat{x}(t),u(t)} \quad H(t) = \frac{\partial h}{\partial x}\bigg|_{\hat{x}(t),u(t)}
\end{align*}
The target uncertainty of the robot state is known from the sampling stage. With the uncertainty from sensing calculated, we are able to compute the terminal error covariance of the Kalman filter state $$\hat\Sigma_k(t)=\Sigma(t)-P_k(t)$$ and use it as output state feedback to control the covariance of the path in the next iteration. 

$\hat{A}_k(t),\hat{a}_k(t)$ represent the Gaussian Markov process approximation of the trajectory at the current iteration, which can be calculated by linearizing the system with respect to the nominal trajectory. $\hat{A}_k(t),\hat{a}_k(t)$ are used in the construction of cost matrices $Q_k(t)$ and $r_k(t)$. Solving the linear covariance steering problem in \ref{eq:iterlinear}, the optimal control policy $K_k(t), d_k(t)$ is calculated and $A_{k+1}(t),a_{k+1}(t)$ are updated following \eqref{eq:updateAa}, as shown in Figure \ref{fig:backward}.

\begin{algorithm}[ht]
    \caption{PGCS State Connection Algorithm}
    \label{alg:noncov}
 \begin{algorithmic}[1]
    \STATE Start state and covariance $m_0, \Sigma_0, P_0$
    \STATE Target state and covariance $m_T, \Sigma_T$
    \STATE Initialize $A_0, a_0$
    \FOR{iteration $k = 0,1,2,\ldots$}
    \STATE $ \Bar{x}_k(t),\;\Sigma_k(t)\leftarrow {\rm UpdateTrajectory}(A_k, a_k)$;
    \STATE $P_k(t) \leftarrow {\rm KalmanGain}(\Bar{x}_k(t),P_k(0))$\eqref{eq:KFriccati};
    \STATE Update $\hat{\Sigma}_k = \Sigma(t) - P_k(t)$
    \STATE $\hat{A}_k(t),\hat{a}_k(t) \leftarrow {\rm Linearization}(\Bar{x}_k(t))$;
    \STATE $K_k(t),d_k(t) \leftarrow {\rm PGCS}(m_0,\hat{\Sigma}_0, m_T, \hat{\Sigma}_T)\eqref{eq:iterlinear}$;
    \STATE Update $A_k(t)$ using \eqref{eq:updateAa1};
    \STATE Update $a_k(t)$ using \eqref{eq:updateAa2};
    \ENDFOR
 \end{algorithmic}
\end{algorithm}

\subsection{Entropy regularized edge cost}
For every trajectory between states, the cost is calculated using the sum of control energy, collision cost, and entropy cost. Entropy cost is defined as
\begin{equation*}
\label{eq:entropy_cost}
    E(\Sigma_k^i(t))=-\int_0^T \log(|\Sigma_k^i(t)|)dt.
\end{equation*}
A smaller entropy cost indicates the trajectory allows higher tolerance in the robot uncertainty and requires less sensing and control effort to control the uncertainty. Leveraging the duality between stochastic control and variational inference, the objectives for the linearized system in each step of our edge construction problem formulation \eqref{eq:covcontrol} is equivalent to an entropy-regularized motion planning \cite{yu2023gaussian} \cite{yu2023stochastic}
\begin{equation*}
    {\max}\; \mE_q[-\log J] + H(q),
\end{equation*} 
where $J$ denotes a composite cost involving a prior process-induced cost and the collision cost
\begin{equation*}
    J = {\rm J_{prior}}(X) + V(X),
\end{equation*}
and $q$ is the joint Gaussian distribution induced by the stochastic process \eqref{eq:robotdynamic} after linearization. In other words, optimizing the problem \eqref{eq:covcontrol} is equivalently optimizing an entropy-regularized motion planning objective for the path distribution. We found that a trajectory distribution with a smaller entropy cost is safer than one with a higher cost in a probability sense. In the same spirit, we define the total cost for the $\textit{i}^{\rm th}$ trajectory $z_k^i$ is the weighted sum of the control energy along the mean trajectory and the entropy cost
\begin{equation}\label{eq:cost}
    c_{ij} = \frac{1}{2}\int_0^T||u^*(t)||^2 + \lVert {\rm hinge}(S(X(t)) \rVert^2 dt + \alpha E(\Sigma_k^i(t)).
\end{equation}
By setting $\alpha$ differently, the planner can return different optimal paths with lower control effort or lower risks.

\subsection{Uncertainty-aware State Sampler}
We utilize BRM to divide the original problem into several easier state connection subproblems.
To leverage the PGCS state connection Algorithm \ref{alg:noncov}, it is important to provide a meaningful covariance to represent the uncertainty for each sample state. Define the distance $d_{obs}$ between an obstacle region $\cX_{obs}$ and sampled state $x_s$ as the minimum distance from $x$ to any point $p_{obs} \in \cX_{obs}$, and the corresponding point in $\cX_{obs}$ is the closet point $p_{obs}^c$ to $x$. For $n$ dimensional spatial state space, 
we hope to find $n$ such points $p_{obs}^c$ and form a covariance ellipsoid with the center point $x_s$. The covariance for spatial states can be calculated from the parameter for this ellipsoid and a given confidence level $P_{conf}$, such that the actual state $x$ distribution satisfies
\begin{equation}
    P((x-x_s) < d_{obs}) > P_{conf}.
\end{equation}
We assume a constant velocity and covariance at each sampled state. The direction of the velocity can align with the direction of the current node and adjacent node.

\subsection{Main algorithm}
The implementation of the PGCS-BRM algorithm is summarized in Algorithm \ref{alg:mp}. To calculate the hinge loss of obstacles, a signed distance field is used which, together with the start, and target states, are initialized by the user. Then the uncertainty-aware sampler samples a certain number of states in the state space and their covariance matrices determined by the environment. The main loop starts in line 11, where each feasible sampled state is looped through and whose nearest neighbors are found. The number of neighbors found is determined by a preset neighbor distance and the total number of sampled states. Next, we connect the current state with all its feasible neighbors using the state connection Algorithm \ref{alg:noncov}. For each state pair, the nonlinear covariance steering connection algorithm is run twice to generate two trajectories from two different directions. To ensure the connection algorithm returns a feasible solution, we need the estimated robot state error-covariance $\hat{\Sigma} > 0$. 

\begin{algorithm}[ht]
    \caption{PGCS-BRM}
    \label{alg:mp}
 \begin{algorithmic}[1]
    \STATE Initialize map $M$;
    \STATE Initialize graph $G \leftarrow \emptyset$;
    \STATE Start state and covariance $S\leftarrow{x_{init}, \Sigma_{init}}$;
    \STATE Target state and covariance $T\leftarrow{x_{target}, \Sigma_{target}}$;
    \FOR{$i = 1,2,\ldots$}
        \STATE $b_i=(x_i,\Sigma_i) \leftarrow {\rm UncertaintyAwareSampler(M)}$;
        \IF{$\rm Feasible(b_i)$}
            \STATE $G \leftarrow b_i$;
        \ENDIF
    \ENDFOR
    \FOR{$i = 1,2,\ldots$}
        \STATE $\{b_j\}_i \leftarrow \rm NearestNeighbors(b_i,M,G)$;
        \FOR{$j = 1,2,\ldots$}
        \STATE $\Bar{x}_{ij}(t),\;\Sigma_{ij}(t) \leftarrow \rm StateConnection(b_i, b_j)$ \ref{alg:noncov};
        \IF {$\rm CollisionFree(\Bar{x}_{ij})$}
            \STATE $c_{ij} \leftarrow \rm Cost(\Bar{x}_{ij}(t),\;\Sigma_{ij}(t))$;
            \STATE $G \leftarrow (\Bar{x}_{ij}(t),\;\Sigma_{ij}(t),c_{ij})$
        \ENDIF
        \ENDFOR
     \ENDFOR
     \STATE $\Bar{x}_{ST}(t),\;\Sigma_{ST}(t) \leftarrow \rm SearchPath(S,T,G)$
     \RETURN Belief Roadmap G, Path ($\Bar{x}_{ST}(t),\;\Sigma_{ST}(t))$
 \end{algorithmic}
\end{algorithm}

\begin{figure*}[t!]
\begin{subfigure}[b]{0.33\textwidth}
    \centering
  \includegraphics[width=\linewidth]{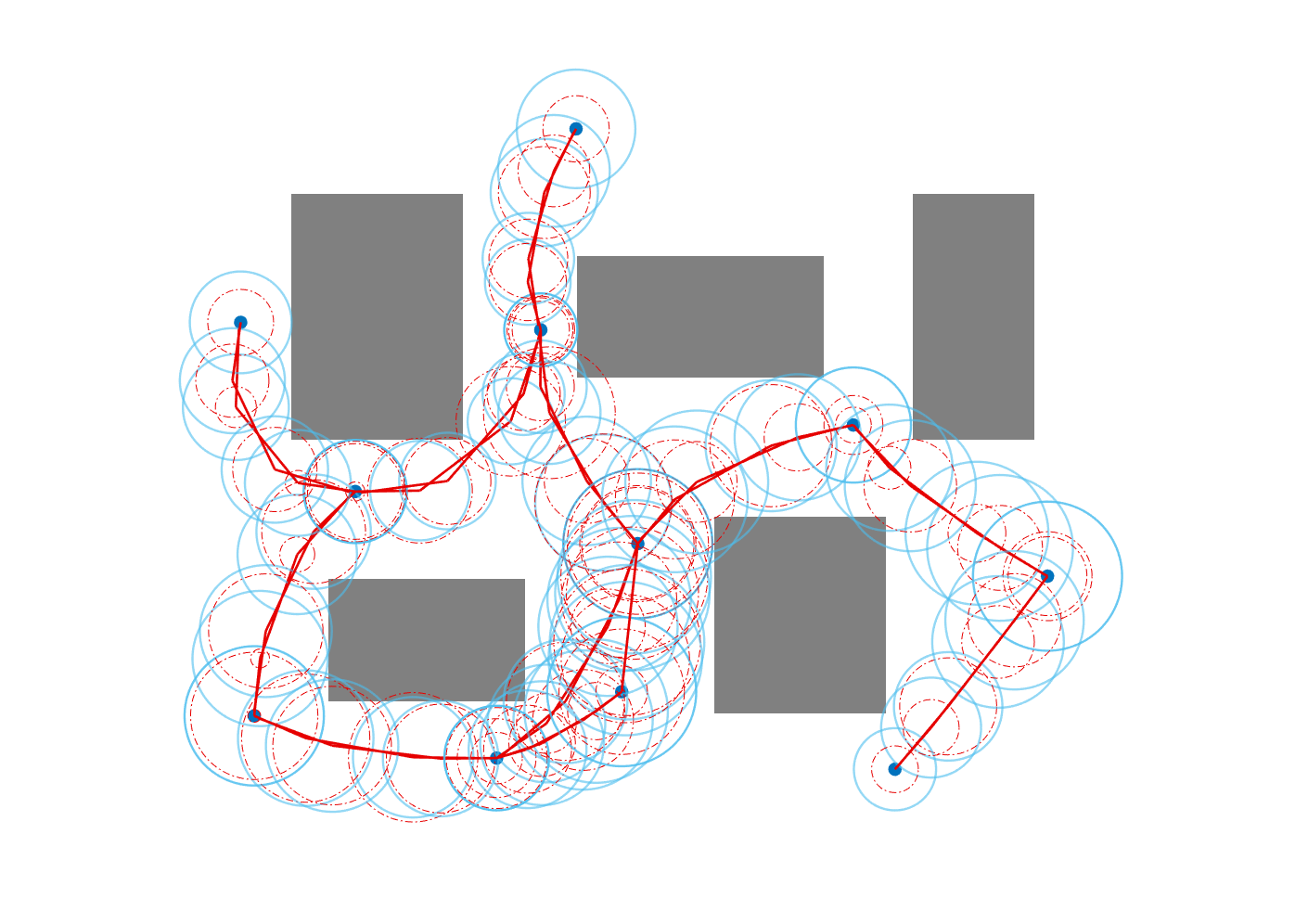}
  \caption{10 nodes and 24 edges}
  \label{fig:fir-st}
\end{subfigure}%
\hfill 
\begin{subfigure}[b]{0.33\textwidth}
    \centering
  \includegraphics[width=\linewidth]{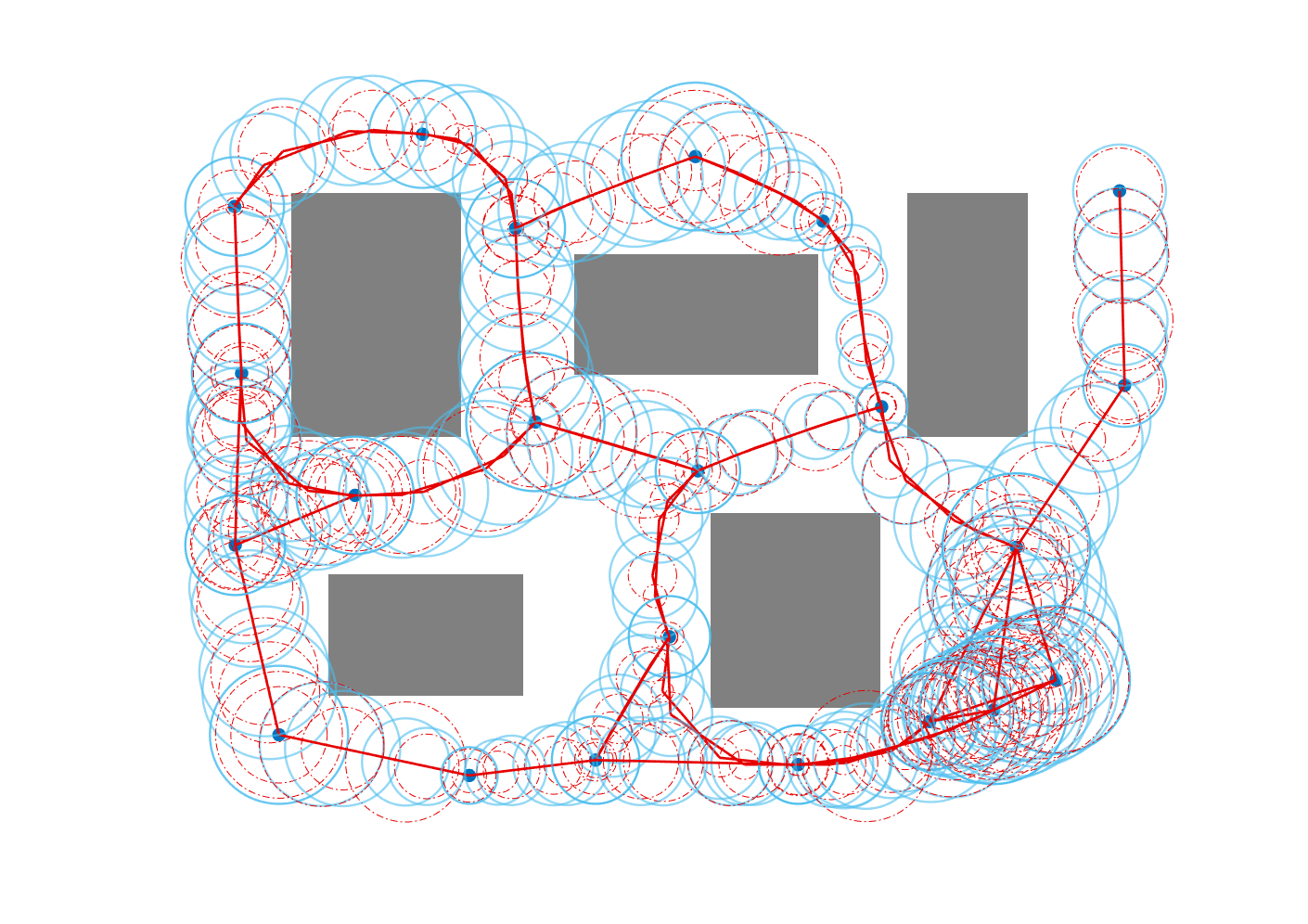}
  \caption{20 nodes and 62 edges }
  \label{fig:second}
\end{subfigure}%
\hfill
\begin{subfigure}[b]{0.33\textwidth}
    \centering
  \includegraphics[width=\linewidth]{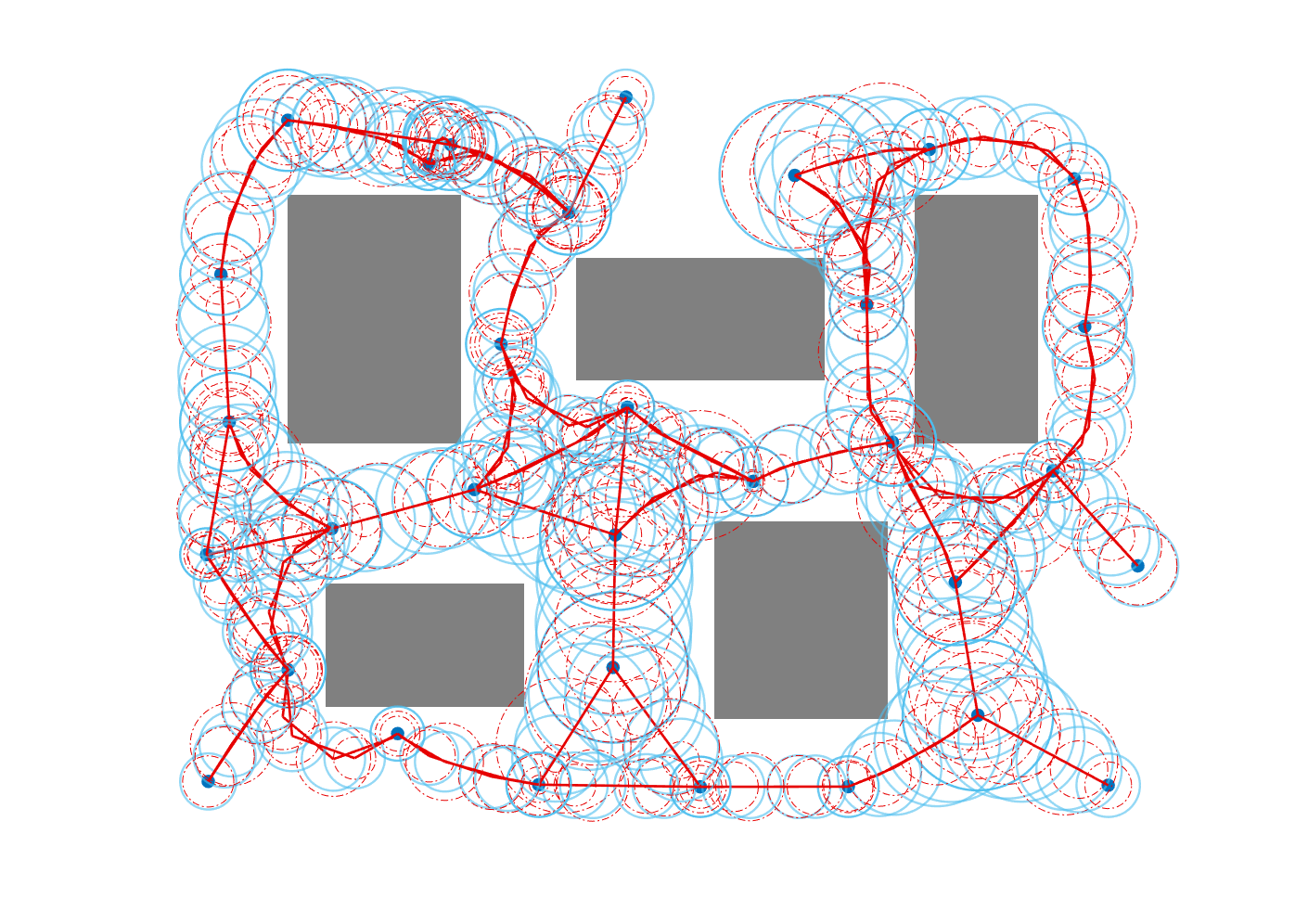}
  \caption{30 nodes and 92 edges}
  \label{fig:third}
\end{subfigure}%
\caption{Belief Roadmap planning for a 2D environment. Red dashed ellipsoids represents the estimated state covariances $P(t)$ propagated using \eqref{eq:KFriccati}, and light blue ellipsoids are the state covariances $\Sigma(t)$. Notice that $P_T$ is expected to be less than $\Sigma_T$ at every end of an edge.
}
\end{figure*}

\begin{table*}[t!]
\centering
\begin{tabular}{|c|cc|cc|cc|cc|}
\hline
Nodes  & \multicolumn{2}{c|}{25}            & \multicolumn{2}{c|}{50}            & \multicolumn{2}{c|}{75}              & \multicolumn{2}{c|}{100}             \\ \hline
Edges  & \multicolumn{1}{c|}{106}   & 104   & \multicolumn{1}{c|}{302}   & 310   & \multicolumn{1}{c|}{476}    & 588    & \multicolumn{1}{c|}{940}    & 974    \\ \hline
Time/s & \multicolumn{1}{c|}{45.41} & 23.77 & \multicolumn{1}{c|}{75.74} & 91.82 & \multicolumn{1}{c|}{111.25} & 282.64 & \multicolumn{1}{c|}{341.83} & 227.62 \\ \hline
\end{tabular}
\caption{Time consumption in different graph scales for a robot in 3D environment. We conduct two different experiments for the same number of nodes to show that when the number of nodes are relatively small, the variance in graph construction time is large. 
} 
\label{tab:time}
\end{table*}

\section{Experiment}
We conducted several numerical experiments to validate the proposed method. All experiments are conducted on a machine with CPU of i7-12700KF and 16GiB memory. 

\subsection{Effect on Changing $\alpha$}
\begin{figure*}[t!]
\begin{subfigure}[b]{0.3\textwidth}
    \centering
  \includegraphics[width=\linewidth]{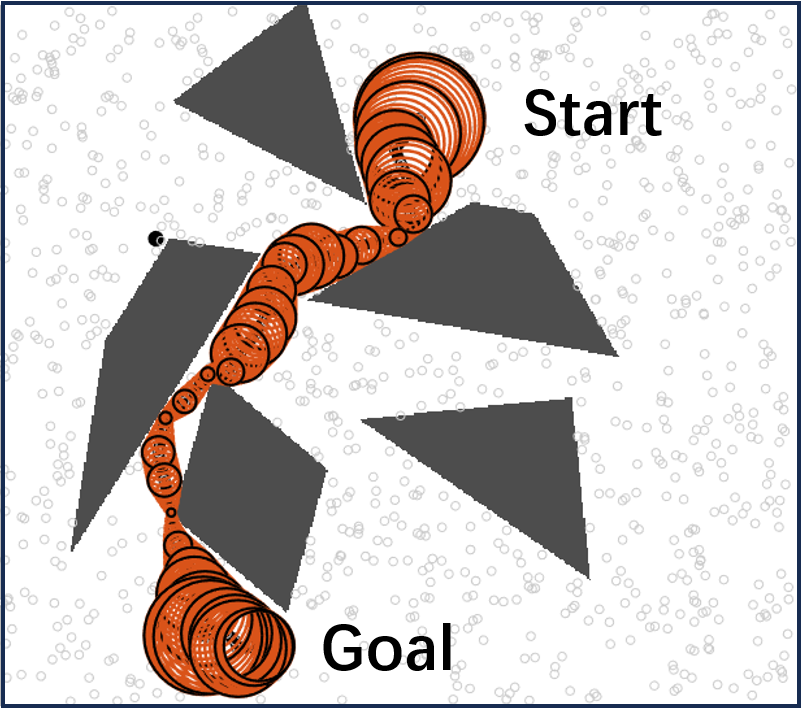}
  \caption{$\alpha=0.2$ Penalize control energy cost more than entropy cost. PGCS-BRM returns a shorter path but more risky path. }
  \label{fig:first_alpha}
\end{subfigure}%
\hfill 
\begin{subfigure}[b]{0.3\textwidth}
    \centering
  \includegraphics[width=\linewidth]{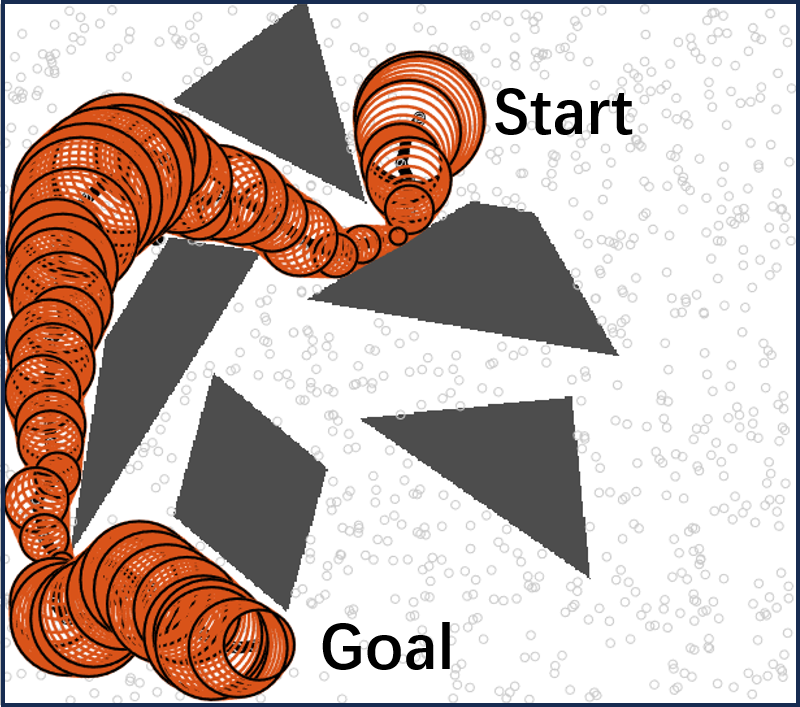}
  \caption{$\alpha=0.6$ Penalize entropy cost more than control energy cost. PGCS-BRM returns a longer but less risky path.}
  \label{fig:second_alpha}
\end{subfigure}%
\hfill
\begin{subfigure}[b]{0.34\textwidth}
    \centering
  \includegraphics[width=\linewidth]{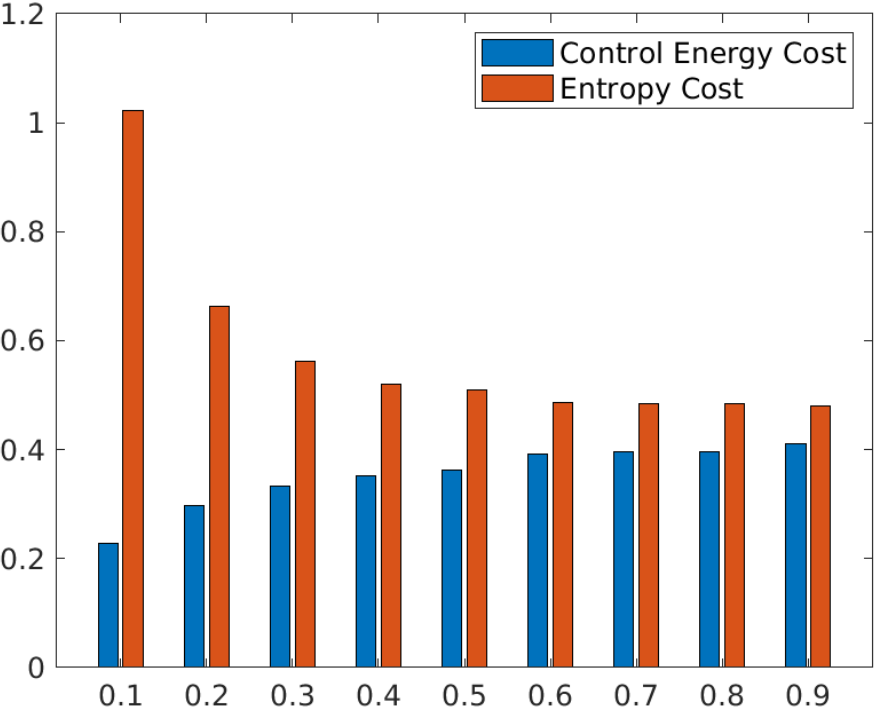}
  \caption{Control energy cost and entropy cost with different $\alpha$. With an increasing $\alpha$, the entropy wins over the control energy costs.}
  \label{fig:cost-comp}
\end{subfigure}%
\caption{Different paths are chosen by different weights on entropy.}
\end{figure*}
To demonstrate how $\alpha$ impacts the returned path and the ability of PGCS-BRM to handle non-linear systems, we conduct experiments on 2-D planning for a risky area. We consider the same nonlinear dynamical system used in \cite{RidOkaTsi19}
    \begin{subequations}\label{eq:double_integrator}
    	\begin{align}   
    			d x_1&=x_2 d t,\\
    			d x_2&=(u - c_d \lVert x_2 \rVert x_2) d t+ \sqrt{\epsilon} d W_t.
    	\end{align}
    \end{subequations}
1,000 states are sampled and more than 10,000 trajectories are generated for state connection. In the search phase, $A^*$, a best-first search algorithm, is deployed to find a path to the given goal state with minimum total cost. In Figure \ref{fig:first_alpha} and \ref{fig:second_alpha}, we show that by changing the weighting factor $\alpha$, PGCS-BRM is able to build belief graphs and find a path with less control cost or less entropy cost.

\subsection{Evaluation of Running Time}
We compare the proposed method with the CS-BRM method in \cite{zheng_belief_2021} using a linear double integrator dynamics
\begin{equation}
\label{eq:linear_dyn}
    dX_t = 
    (\begin{bmatrix}
    0 & I\\
    0 & 0
    \end{bmatrix} X_t + \begin{bmatrix}
        0 \\ I
    \end{bmatrix} U) d t + \sqrt{\epsilon} dW_t.
\end{equation}
We use a map of 5 rectangular obstacles to compare these two methods. In each experiment with a different number of nodes, the same sampling setup is deployed and we used the same start and goal states for the graph building. For PGCS-BRM, each edge building is set to execute 50 iterations of the proximal gradient with step size $\eta=0.001$ and discretized into 50 timesteps. We recorded the times for constructing the belief space graph after node sampling and repeated each experiment three times. Both algorithms are able to build a belief roadmap, however, due to the high computation cost in performing Monte-Carlo collision checking and solving optimization problems, CS-BRM requires higher computation time. On the other hand, PGCS-BRM is able to penalize collision in the cost function and directly solve the nonlinear covariance steering with a sublinear rate. PGCS-BRM is around 100 times faster and only requires 3.38s to build a roadmap with 30 nodes, compared to CS-BRM which needs 782s on average.

\begin{figure}[h] 
    \centering
    \includegraphics[width=0.5\textwidth]{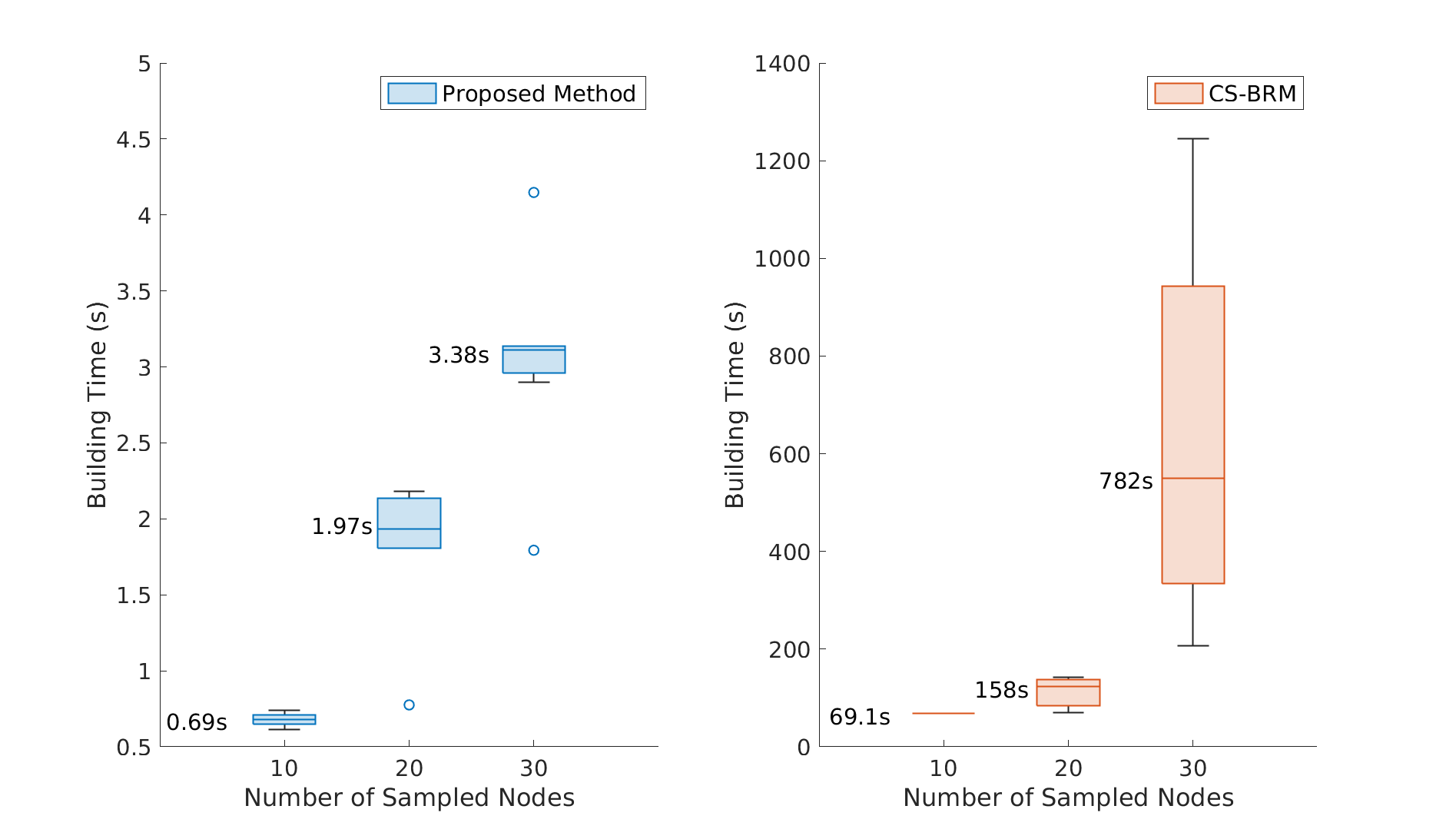}
    \caption{Running time comparison of graph construction between PGCS-BRM and CS-BRM \cite{zheng_belief_2021}. The proposed method is more than \textbf{100 times more efficient} in graph building.}
    \label{fig:compare_boxplot}
\end{figure}

\subsection{3D Experiment}

\begin{figure}[h] 
    \centering
    \includegraphics[width=0.4\textwidth]{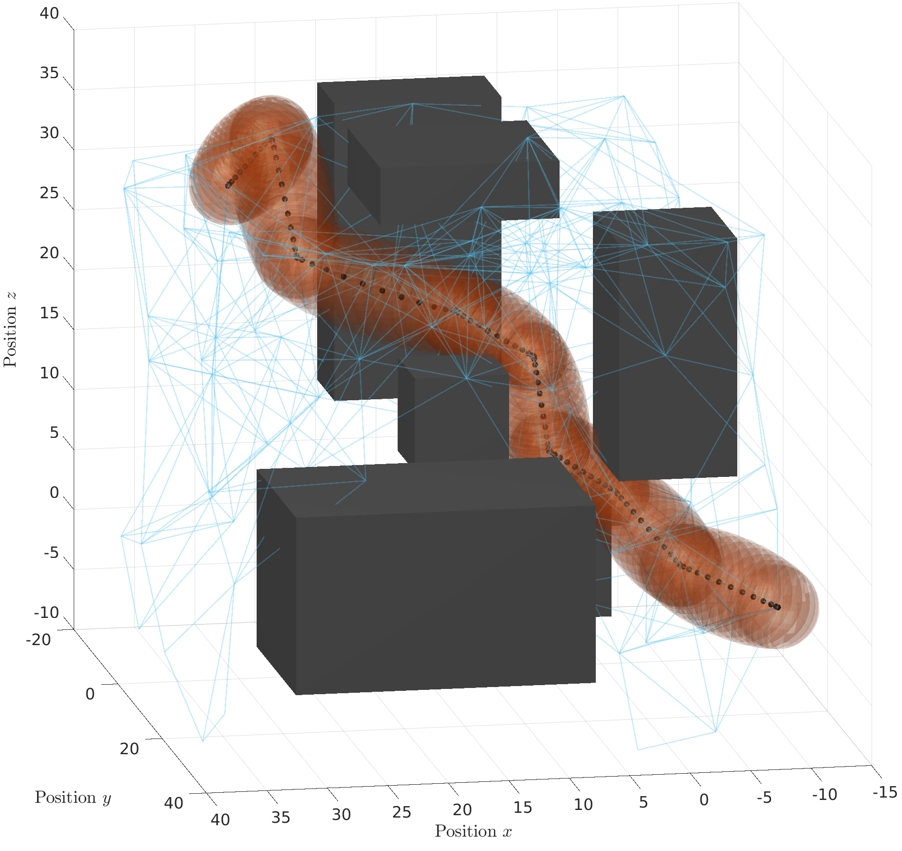}\label{fig:sampler}
    \caption{Belief space graph and an optimal path in 3D space. The graph consists of 100 sampled states. The red funnel shows the optimal path found by the PGCS-BRM.}
    \label{fig:3D-PGCSBRM}
\end{figure}

We conduct experiments in an obstacle-clustered 3-D environment for a 3-D point robot model \eqref{eq:linear_dyn} to demonstrate the generalizability of the proposed methods in higher-dimension space. The visualization result is shown in Figure \ref{fig:3D-PGCSBRM}. We record the average time for building a PGCS-BRM with 25, 50, 75, and 100 nodes and different numbers of edges. On average, PGCS takes 0.23-0.48s to build an edge in 3D space under 100 max iterations. It is worth noticing that the actual run times vary because different states are sampled in each experiment.

\section{Conclusion and future work}\label{sec:conclusion}
This work presents an efficient belief space roadmap (PGCS-BRM) for planning under uncertainty. The proposed method models the belief as state distributions and leverages nonlinear covariance steering with safety constraints for edge construction. We also include an entropy cost in the edge costs to account for robustness under uncertainty. Experiments show that the proposed method effectively constructs BRMs in different dimensions and outperforms state-of-the-art sampling-based belief space planning methods. Though PGCS-BRM shows promising results in building a belief roadmap with controlled covariance, the generated trajectory is still rough and not smooth. This is mainly the result of the lack of reasonable velocity sampling. Unlike spatial states, there are no explicit constraints on velocity in the sampling stage, and poorly selected velocity might result in a non-smooth trajectory. Developing a better velocity sampling algorithm and smoothing algorithm can greatly enhance the performance of the current algorithms.
Another future direction worth exploring is to deploy such an algorithm in time-varying environments. The ability to control the uncertainty in planning and quickly replan the route is essential in such scenarios.

\newpage
{
\bibliographystyle{IEEEtran}
\bibliography{./refs}
}
\end{document}